\documentclass{article} 
\usepackage{iclr2027_conference,times}

\usepackage{amsmath,amsfonts,bm}

\def\eqref#1{equation~\ref{#1}}

\def\1{\bm{1}}

\DeclareMathAlphabet{\mathsfit}{\encodingdefault}{\sfdefault}{m}{sl}
\SetMathAlphabet{\mathsfit}{bold}{\encodingdefault}{\sfdefault}{bx}{n}

\usepackage{hyperref}
\usepackage{url}
\usepackage{enumitem}

\newcommand{\paratitle}[1]{\vspace{1.5ex}\noindent\textbf{#1}}

\newcommand{\ignore}[1]{}

\usepackage{booktabs}
\usepackage{tcolorbox}
\usepackage{wrapfig}
\usepackage{caption}
\usepackage{multirow}

\title{Less Is More: Learning What to Profile for GPU Kernel Optimization}

\author{
Yihan Wang \\
Renmin University of China \\
\texttt{yihan3123@gmail.com}
\And
Peiyu Liu\thanks{Corresponding author.} \\
University of International Business and Economics \\
\texttt{liupeiyustu@163.com}
\And
Runyu Chen \\
University of International Business and Economics \\
\texttt{ry.chen@uibe.edu.cn}
\And
Wei Xu \\
Renmin University of China \\
\texttt{weixu@ruc.edu.cn}
}
\iclrfinalcopy 
\begin{document}

\maketitle

\begin{abstract}
Large language models have recently shown strong potential for automatically optimizing GPU kernels, yet their effectiveness depends critically on the hardware information available to the kernel optimization model. 
Existing systems typically rely on predefined profiling strategies, leaving \emph{what to profile} largely outside the optimization process.
We show that effective profiling is not simply about collecting more information: increasing the number of profiling metrics does not monotonically improve kernel optimization, while the most useful profiling information varies across kernels. Our theoretical analysis further shows that, under such heterogeneous profiling utility, kernel-conditioned selection can outperform any single fixed profiling configuration.
Motivated by this observation, we propose \emph{AutoProfiler}, a feedback-driven framework for learning a deterministic profiling tree from optimization feedback.
The tree is iteratively refined through REWRITE and SPLIT operations, with an LLM generating candidate modifications and empirical evaluation determining their acceptance. 
Once constructed, the tree quickly gathers the required profiling information for any kernel, allowing the LLM rewriter to be invoked only once.
Experiments on KernelBench show that our approach improves kernel optimization over fixed profiling configurations under constrained profiling budgets. 
These results suggest that hardware profiling should evolve from static metric collection toward profiling reasoning, making profiling itself a new dimension for improving kernel optimization. The codes are available at~\href{https://github.com/Satissss/AutoProfiler}{https://github.com/Satissss/AutoProfiler}.
\end{abstract}


\section{Introduction}
LLM-based systems have recently shown strong potential for automatically rewriting GPU kernels to improve execution efficiency~\cite{ouyang2025kernelbench,zhang2025cudaforge,gai2026kernelpro}. 
To better guide the rewriting process, recent systems increasingly incorporate hardware profiling to expose runtime bottlenecks and provide actionable optimization feedback~\cite{zhang2025cudaforge, gai2026kernelpro}.
However, existing approaches typically collect a predefined set of profiling signals before optimization.
While simple and reusable, this design overlooks two fundamental properties of hardware profiling: 
profiling information is costly to acquire, and its utility is inherently kernel-dependent.

This raises a fundamental question: 
\emph{What profiling information is actually needed for effective kernel optimization?}
We investigate this question from two complementary perspectives. First, more profiling is not necessarily better. Empirically, increasing the number of profiling metrics does not monotonically improve kernel optimization, suggesting that more profiling information does not necessarily translate into better optimization. 
Second, the useful information is kernel-dependent. Our theoretical analysis shows that when profiling utility varies across kernels, kernel-conditioned selection can outperform any single fixed configuration, which is further supported by our empirical oracle analysis. Together, these findings suggest that effective profiling is not about collecting more information, but about identifying the right information for each kernel.

{To this end, we propose AutoProfiler, which learns a kernel-conditioned profiling policy that adaptively acquires profiling metrics before invoking the kernel rewriter.}
We formulate profiling as sequential information acquisition and represent the policy with a deterministic profiling tree, which selects subsequent measurements based on previously observed profiling results. The tree is refined through optimization feedback: an LLM proposes structural REWRITE to deepen the tree by collecting more profiling information, and SPLIT to widen the tree for kernels with different profiling needs, 
while empirical paired evaluation determines whether a proposed modification is accepted. The LLM is used only during policy construction, leaving inference-time routing fully deterministic.

Experiments on KernelBench show that AutoProfiler improves kernel optimization over fixed profiling configurations under a constrained profiling budget, while recovering a substantial portion of the gain available from kernel-conditioned selection.
Interestingly, the learned profiling policy remains effective across iterative optimization rounds, enabling continued gains without repeated LLM-based profiling decisions.
Together, these results point to a promising direction for self-improving kernel generation: using more targeted profiling to provide effective feedback at lower cost, thereby supporting continued improvement through iterative optimization.
\begin{figure}[t]
    \centering
    \includegraphics[width=0.8\linewidth]{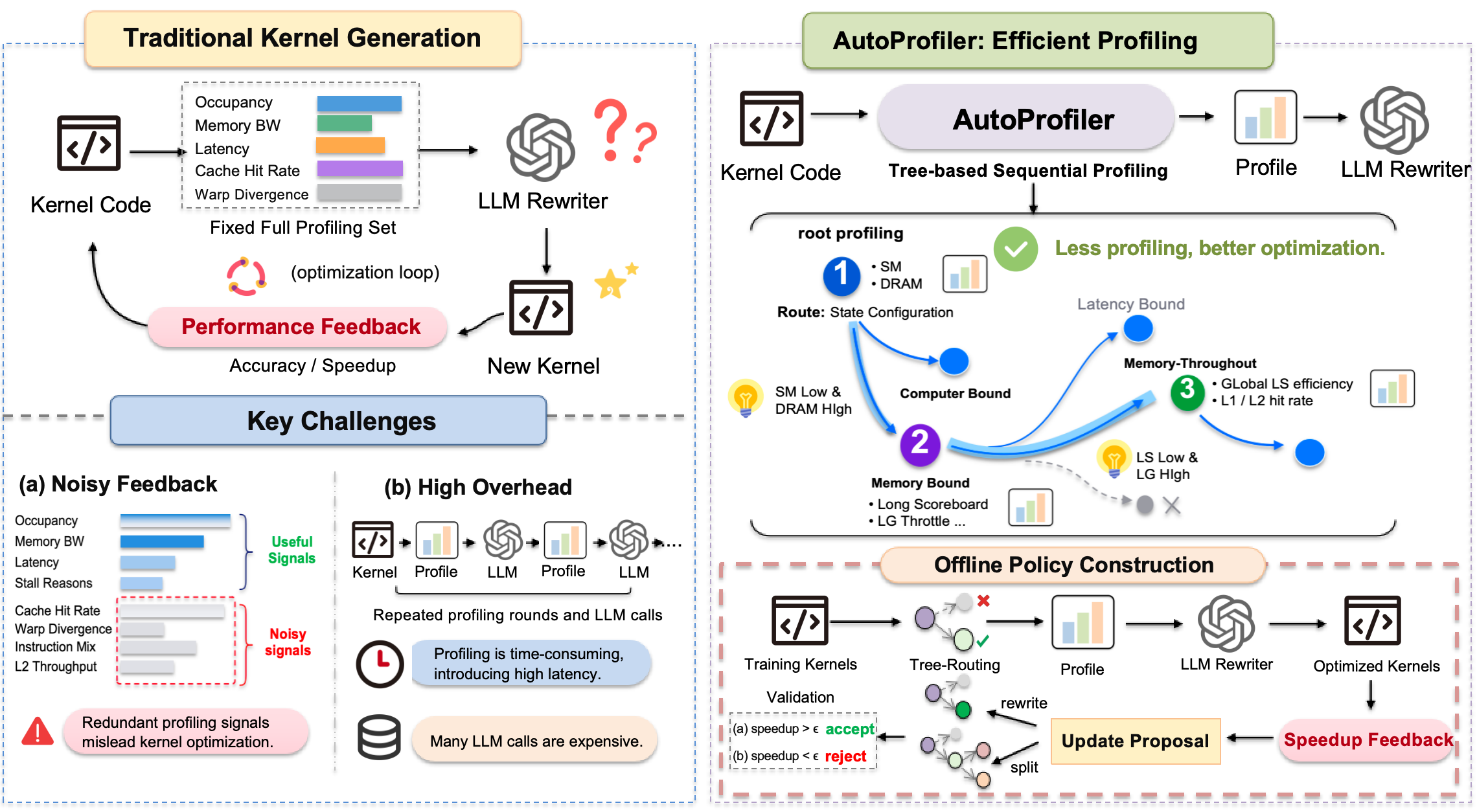}
    \caption{Overview and motivation of our profiling reasoning framework. \textbf{Left:} Traditional kernel optimization and its key challenges, where redundant profiling signals introduce noisy feedback (a) and repeated profiling rounds incur high latency and LLM overhead (b). \textbf{Right:} Our AutoProfiler method utilizes tree-based sequential profiling to selectively acquire informative signals, achieving effective optimization with reduced profiling overhead.}
    \label{fig:pareto}
\end{figure}

\section{Preliminary}\label{preliminary}
\paragraph{Notations.} Let $k\in\mathcal{K}$ denote a GPU kernel sampled from a kernel corpus, and let $\mathcal{M}$ denote the set of available profiling metrics. Given a subset of metrics $m\subseteq\mathcal{M}$, a profiler $P$ produces the corresponding profiling information $I=P(k,m).$ A kernel rewriter $g$ then uses the original kernel and profiling information to generate an optimized kernel: $\hat{k}=g(k,I)=g\big(k,P(k,m)\big).$ Throughout this work, the profiler $P$, the rewriter $g$, and the target hardware are fixed. 

\paragraph{Problem setup.} 
We formulate the GPU kernel optimization as a profiling-conditioned generation problem: given an input kernel, the optimized kernel is generated conditioned on both the kernel itself and the acquired hardware profiling information. 
Existing approaches typically use a fixed profiling configuration for all kernels. Given a predefined metric set $m_0\subseteq\mathcal{M}$, the optimized kernel is generated as $\hat{k} = g\bigl(k,P(k,m_0)\bigr)$, where the same set of profiling metrics is collected regardless of the kernel being optimized.
In contrast, we allow the profiling configuration to vary across kernels. Let $\phi(k)\subseteq\mathcal{M}$ denote the profiling metrics selected for kernel $k$. The optimized kernel is then generated as 
\begin{equation}
    \hat{k} = g\bigl(k,P(k,\phi(k))\bigr).
\end{equation}
This formulation turns profiling into a sequential reasoning process over hardware signals: based on the profiling information observed so far, the policy determines what additional information is most valuable to acquire next for kernel optimization.

\paragraph{Our goal.} 
We aim to learn a profiling policy $\phi$ that determines what profiling information to acquire so as to improve kernel optimization. We measure the utility of the acquired profiling information by the performance of the resulting kernel. Specifically, for a kernel $k$ and profiling metrics $m$, we define the resulting speedup as $s(k,m)=\frac{\operatorname{lat}(k)}
{\operatorname{lat}(\hat{k})}$, where $\operatorname{lat}(\cdot)$ denotes the execution latency on the target hardware. The corresponding optimization reward is $r(k,m)=\max\left\{s(k,m)-1,0\right\}$, which assigns positive reward only when the acquired profiling information leads to a rewrite that improves upon the original kernel. Our objective is therefore to learn a profiling policy that maximizes the expected optimization reward:
\begin{equation}
    \arg\max_{\phi}
\mathbb{E}_{k\sim\mathcal{K}}
\left[
r\big(k,\phi(k)\big)
\right],
\end{equation}
subject to a profiling budget that limits the amount of hardware information acquired for each kernel, where $\mathcal{K}$ denotes the distribution over kernels under consideration. The key challenge is therefore to exploit kernel-dependent profiling utility while avoiding unnecessary profiling cost.

\section{Static vs. Dynamic Metric Selection: Theoretical and Empirical Analysis}\label{analysis}

\paratitle{Theoretical Analysis.} We first characterize the fundamental advantage of dynamic profiling over static metric selection. Specifically, we show that dynamic selection can never perform worse than the best fixed profiling configuration, and that a strict advantage arises when different configurations are effective for different kernels. Consider a finite family of candidate profiling configurations $\mathcal{C} = \{m_1, m_2, \dots, m_K\}$ and a kernel distribution $\mathcal{D}$. Let $u(k, m)$ denote the kernel optimization utility (\emph{e.g.,} performance speedup) for kernel $k$ under configuration $m$. The expected utility of the optimal static configuration is defined as $U_{\mathrm{static}} = \max_{m \in \mathcal{C}} \mathbb{E}_{k \sim \mathcal{D}}[u(k, m)]$, whereas an oracle dynamic policy that adaptively selects the most effective configuration per kernel achieves $U_{\mathrm{dynamic}} = \mathbb{E}_{k \sim \mathcal{D}}[\max_{m \in \mathcal{C}} u(k, m)]$.

\paratitle{Proposition 1. Dynamic selection dominates static selection.} For any finite candidate family $\mathcal{C}$,
\begin{equation}
U_{\mathrm{dynamic}} \geq U_{\mathrm{static}}.
\end{equation}

\begin{wrapfigure}{r}{0.50\textwidth}
    \centering
    \includegraphics[width=0.46\textwidth]{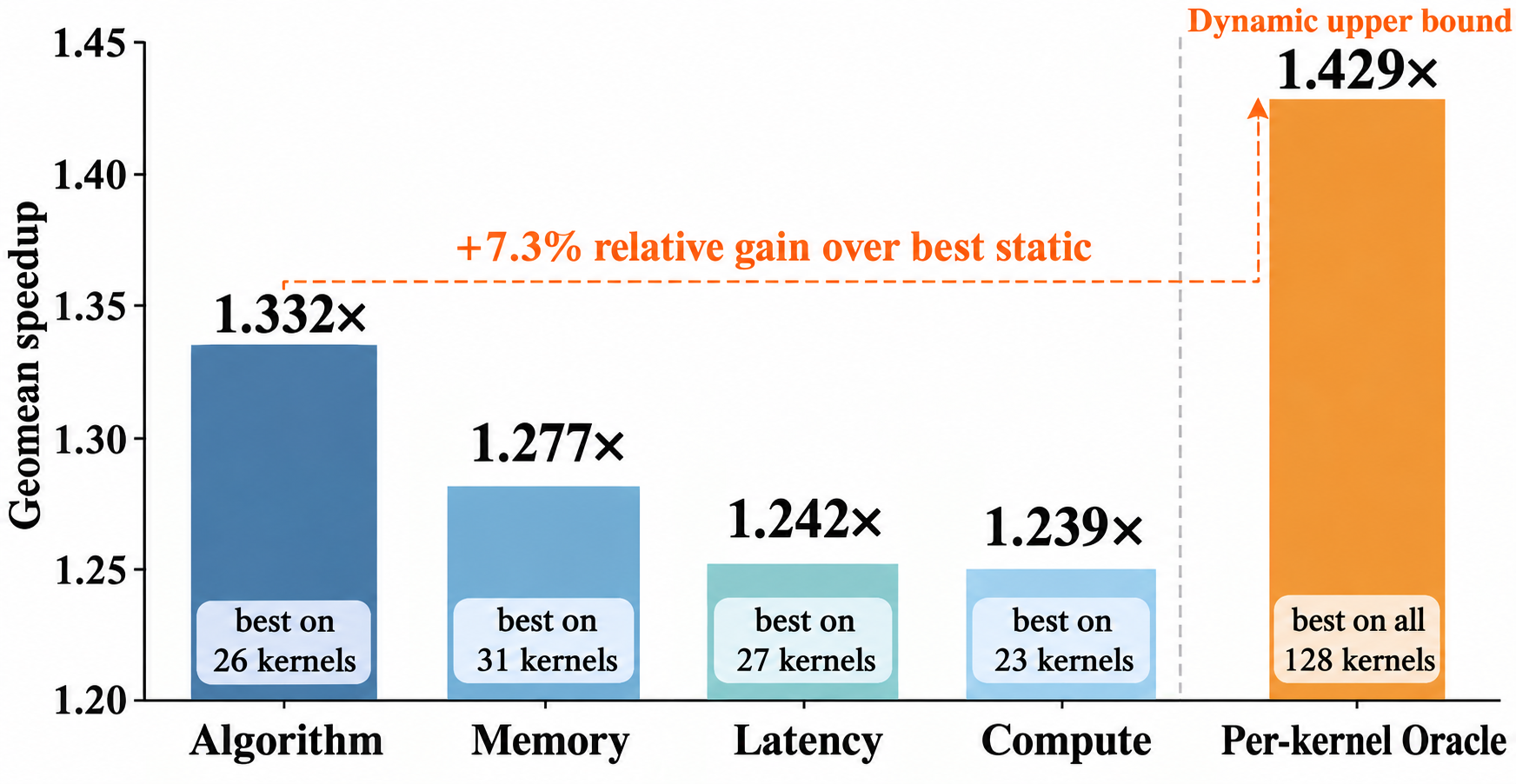}
    \caption{Per-kernel oracle evaluation on 128 KernelBench kernels, showing substantial heterogeneity in the optimal profiling configuration and a $7.3\%$ relative gain over the best fixed configuration.}
    \label{fig:pareto}
\end{wrapfigure}

Moreover, equality holds if and only if there exists a configuration $m_j \in \mathcal{C}$ whose success region covers the union of all candidate success regions almost surely under $\mathcal{D}$. Otherwise, $U_{\mathrm{dynamic}} > U_{\mathrm{static}}$. This result shows that the value of dynamic profiling is determined by the heterogeneity of profiling utility across kernels: when no single configuration covers the success regions of the others, adapting the profiling configuration to each kernel yields a strict advantage.

\paratitle{Empirical Validation.} We next examine whether the heterogeneity underlying Proposition 1 occurs in practice. We conduct an oracle-style evaluation on 128 KernelBench kernels using four roofline-based profiling configurations. For each kernel, we evaluate the same frozen optimizer under each configuration and select the configuration yielding the highest speedup. Detailed experimental settings are provided in Appendix~\ref{app:empirical}.
We find that the per-kernel oracle achieves a geomean speedup of $1.429\times$, compared with $1.332\times$ for the best fixed configuration, corresponding to a $7.3\%$ relative gain. Moreover, the oracle selects different configurations across kernels, with the best configuration varying substantially among the four candidates. 

\begin{tcolorbox}[colback=blue!5!white,colframe=blue!55!black,width=1.0\textwidth,title={Key Findings on Profiling Utility}]
{
\begin{itemize}[
    leftmargin=1.4em,
    labelsep=0.45em,
    itemsep=0.9ex,
    topsep=0.2ex,
    parsep=0pt
]
\item \textbf{Profiling utility is heterogeneous across kernels.}\\[-0.2ex]
\textit{Different kernels benefit from different profiling signals, and no single configuration is consistently optimal.}

\item \textbf{Kernel-specific profiling provides substantial optimization headroom.}\\[-0.2ex]
\textit{The oracle gap shows that effective optimization depends on selecting useful profiling information for each kernel rather than collecting a universally strong metric set.}
\end{itemize}
}
\end{tcolorbox}

This result provides empirical support for the heterogeneity condition in Proposition 1 and indicates non-trivial room for kernel-conditioned profiling. The oracle serves only as an analysis reference; it requires evaluating all candidate configurations for each kernel and is unavailable at inference time. This motivates the central problem addressed next: learning which profiling configuration to use for a given kernel without exhaustively evaluating all candidates.

\section{Methods}

To optimize each kernel with the most informative profiling feedback, we aim to identify what information is most useful at each stage of optimization.
{In this section, we present AutoProfiler, a feedback-driven framework for learning kernel-conditioned hardware profiling policies. AutoProfiler formulates hardware profiling as a sequential decision process and instantiates the learned policy with a deterministic profiling tree. Our methodology is structured as follows: we first introduce the Tree-based Profiling Policy formulation (Section~\ref{sec:tree}), detail its Feedback-driven Policy Construction via LLM-based structural updates and describe its Deterministic Inference-time Routing pipeline~(Section~\ref{sub-construction}).} 

\subsection{Tree-Based Profiling Policy}\label{sec:tree}
\label{sub-tree}
AutoProfiler model profiling as a sequential decision process and implement the resulting policy with a deterministic tree. We first formalize how profiling information is acquired sequentially, then describe how this policy is represented by the tree, and finally explain how the learned tree guides profiling for an input kernel.

\paratitle{Sequential Profiling.} 
To formalize this process, we formulate profiling as a sequential policy that determines what hardware information to collect next based on the profiling observations obtained so far. For a kernel $k$, at profiling step $t$, the current state is denoted by
\[
s_t=(M_t,I_t),
\]
where $M_t\subseteq\mathcal{M}$ is the set of metrics collected so far and $I_t=P(k,M_t)$ is the corresponding profiling information. The policy then selects an additional metric group
\[
G_t=\pi_T(s_t), \qquad G_t\subseteq \mathcal{M}\setminus M_t,
\]
which leads to the next profiling state
\[
s_{t+1}=(M_{t+1},I_{t+1}),
\qquad
M_{t+1}=M_t\cup G_t,\quad
I_{t+1}=P(k,M_{t+1}).
\]
Equivalently, the profiling process can be viewed as
\[
s_t \xrightarrow{\ \pi_T\ } G_t \xrightarrow{\ \text{collect } G_t\ } s_{t+1}.
\]
Profiling terminates when the policy determines that no additional information is required. In this way, the metrics collected for a kernel depend on its observed hardware behavior rather than being fixed in advance.

\paratitle{Tree Representation.} 
To instantiate this sequential policy, we represent $\pi_T$ as a rooted tree $T=(V,E,v_0)$. 
{Specifically, the root node $v_0$ captures an initial profiling state using two fundamental microarchitectural signals: SM (Compute) utilization and DRAM (Memory) throughput. Guided by the roofline model, these two metrics classify any input kernel into one of four canonical bottleneck regimes: \emph{Latency-bound}, \emph{Memory-bound}, \emph{Compute-bound}, or \emph{Algorithm-bound}. This taxonomy forms the four initial outgoing edges from $v_0$, establishing a coarse-grained routing foundation before fine-grained metric selection.} 
From this initial routing, the tree progressively grows into two types of nodes with distinct roles:
\begin{itemize}
    \item \textbf{Internal node:} This specifies a routing condition over the profiling information collected so far, with outgoing edges corresponding to different outcomes of the condition.

    \item \textbf{Leaf:} This represents a terminal profiling state and determines the final metric configuration for the kernels reaching it.
\end{itemize}

\paratitle{Tree-Guided Profiling.} 
Given a kernel \(k\), the profiling tree sequentially evaluates its routing
conditions and collects the required metrics until reaching a leaf. This produces
a kernel-specific metric set \(E_T(k)\) and the corresponding profiling information
\[
I_T(k)=P\!\left(k,E_T(k)\right).
\]
Importantly, all profiling decisions are completed by the deterministic tree without
LLM-based routing. Once the complete profiling information has been collected, it is
provided to the LLM rewriter in a single inference:
\[
\hat{k}_T=g\!\left(k,I_T(k)\right).
\]
Thus, the tree determines what hardware evidence to acquire, while the LLM consumes
the complete evidence once to generate the final optimized kernel.

\subsection{Feedback-Driven Policy Construction}
\label{sub-construction}

We next turn to how the profiling tree itself is constructed. The construction proceeds iteratively through three steps: collecting leaf-level optimization feedback, proposing local \textsc{Rewrite} or \textsc{Split} operations, and retaining refinements that improve kernel optimization.

\subsubsection{Leaf-Level Attribution and Structural Updates}

Without loss of generality, we consider the current tree
\(T\) with root \(v_0\) and a leaf \(v_i\). We route \(N\) training kernels
\(\{k_j\}_{j=1}^{N}\) through \(T\), among which \(N_i\) kernels eventually
reach \(v_i\). Following the sequential profiling process defined in
Section~\ref{sub-tree}, each such kernel \(k_j\) arrives at \(v_i\) with a terminal
profiling state
\[
s_{t_i}^{(j)}
=
\left(
M_{t_i}^{(j)},
I_{t_i}^{(j)}
\right),
\]
where \(M_{t_i}^{(j)}\) is the set of metrics collected along its routing path
and \(I_{t_i}^{(j)}=P(k_j,M_{t_i}^{(j)})\) is the corresponding profiling
information. After the collected information is provided to the fixed LLM
rewriter, executing the resulting kernel on the target hardware yields the
optimization reward \(r_T(k_j)\).

Therefore, for the leaf $v_i$ we obtain two forms of local evidence: their terminal profiling observations \(I_{t_i}^{(j)}\), which characterize the observed hardware behavior, and their optimization rewards \(r_T(k_j)\), which reflect the resulting optimization performance.  Based on this local evidence, we consider two complementary forms of tree growth : \textsc{Rewrite} enriches the profiling information collected at \(v_i\), while \textsc{Split} further partitions the kernels reaching \(v_i\) into different profiling regimes.

\paratitle{\textsc{Rewrite}.}
One possible limitation of the current leaf is that its profiling configuration
does not expose sufficient information for effective optimization. We therefore
consider augmenting its metric configuration with an additional metric group
\(G\):
\[
M'_v=M_v\cup G,
\qquad
G\subseteq\mathcal{M}\setminus M_v.
\]
\textsc{Rewrite} preserves the routing structure while enriching the information
available to all kernels reaching \(v\).

\paratitle{\textsc{Split}.}
Another possibility is that kernels reaching the same leaf benefit from different
profiling information. We therefore consider partitioning the leaf according to
an already-collected profiling feature \(x\) and threshold \(\theta\):
\[
v\rightarrow
\begin{cases}
v_L, & x\leq\theta,\\
v_R, & x>\theta.
\end{cases}
\]
The resulting child leaves can adopt different profiling configurations, allowing
the policy to specialize subsequent profiling decisions according to the observed
hardware behavior.

For both operators, an LLM uses the local feedback to propose
candidate refinements: additional metric groups for \textsc{Rewrite}, or
feature--threshold pairs for \textsc{Split}. The LLM only generates candidate
modifications; whether a candidate is incorporated into the tree is determined
by empirical evaluation in the subsequent stage.

\subsubsection{Conservative Empirical Acceptance}

A candidate modification is accepted only if it demonstrates sufficient downstream improvement over the incumbent tree. For a candidate tree $T'$, we evaluate $T'$ and the incumbent $T$ on the same set of training kernels and compute their paired reward difference:
$$\frac{1}{N}
\sum_{k=1}^{N}
\left[
r_{T'}(k)-r_T(k)
\right],$$
where $r_{T'}(k)$ and $r_T(k)$ denote the downstream rewards obtained by kernel (k) under the candidate and incumbent trees, respectively. The candidate is accepted only when
$$
\widehat{\Delta}(T',T)>\epsilon,$$
where $\epsilon$ is a conservative acceptance margin.

Paired evaluation controls for kernel-level variation by evaluating the candidate and incumbent trees on the same kernels, isolating the effect of the structural modification. When the rewriter involves stochastic sampling, the same sampling seed is used for the paired evaluations.
We apply the operators conservatively. REWRITE is considered first as the minimal intervention for potential information insufficiency at the current leaf. If no REWRITE candidate satisfies the acceptance criterion, SPLIT is considered to capture potential heterogeneity within that leaf. A rejected modification leaves the incumbent tree unchanged. Together with the profiling budget (B), this controls the complexity of the learned tree and the amount of profiling information collected for each kernel.

Once construction terminates, the resulting tree is fixed and used for inference. The LLM is involved only during offline policy construction; inference-time routing follows the learned profiling conditions deterministically and requires no LLM.
\section{Experiments}
\subsection{Experimental Setup}
\label{sec-setup}
\paratitle{Datasets.}
We use separate datasets for profiling-policy construction and evaluation to avoid overlap between training and test kernels. The profiling tree is constructed solely on the CUDA-Agent-Ops-6K corpus without exposure to KernelBench. As a result, KernelBench serves as an out-of-distribution benchmark to evaluate the policy's transferability to unseen kernels.

For training, we use CUDA-Agent-Ops-6K~\citep{dai2026cudaagent}, a curated dataset of 6,000 synthesized operator-level tasks released for CUDA kernel generation and optimization. The dataset is constructed by collecting reference operators from torch and transformers, synthesizing multi-operator tasks with LLMs, and filtering them based on execution correctness, determinism, runtime, and similarity to KernelBench test cases. This provides a diverse training corpus while reducing contamination with the evaluation benchmark.

For evaluation, we use KernelBench~\citep{ouyang2025kernelbench}, a widely used benchmark for LLM-based GPU kernel optimization. It contains three levels of increasing complexity: Level 1 (100 problems) focuses on individual operators such as matrix multiplication and convolution; Level 2 (100 problems) targets fused operators such as Conv+ReLU+Bias; and Level 3 (50 problems) consists of complete models such as MobileNet, VGG, and MiniGPT. These levels cover a broad range of operator complexity and problem scale. 

\paratitle{Baselines.} We compare AutoProfiler with four training-free agentic kernel-optimization methods: CudaForge~\citep{zhang2025cudaforge}, which iteratively compiles, benchmarks, and repairs generated kernels; KernelMem~\citep{sun2026kernelskill}, which incorporates previous optimization experiences through memory-augmented prompting; KernelAgent~\citep{cheng2026kernelagent}, which searches for improved kernels using correctness and performance feedback; and AutoKernel~\citep{jaber2026autokernel}, which progressively refines kernels through single-edit modifications with rollback. These methods cover different inference-time optimization strategies, providing a diverse set of baselines for evaluating AutoProfiler.

\paratitle{Evaluation Metrics.}
{Following the prior kernel optimization works~\cite{zhang2026kernelbenchverified,ran2025kernelband}, we evaluate model performance using execution accuracy to measure correctness and geometric-mean speedup to measure efficiency. }
Execution accuracy measures the proportion of problems where the generated kernel is both compilable and numerically correct against the PyTorch reference. Geometric-mean speedup is computed relative to the reference kernel across all problems, providing a scale-robust aggregate metric. Unlike the arithmetic mean, which is prone to severe distortion from extreme outliers, the geometric mean ensures that no single task disproportionately dominates the aggregate score, yielding a more balanced and representative evaluation~\citep{fleming1986not}.

\paratitle{Implementations.}
We employ {GPT-5.6-Sol} as the backbone kernel rewriter across AutoProfiler and all baseline methods. To ensure a fair comparison under equal inference budgets, each method is restricted to a single optimization pass ($\text{round}=1$). For hardware profiling, baselines adopt their original static NCU metric configurations, whereas AutoProfiler dynamically selects profiling metrics guided by its learned policy. All methods share an identical NCU backend on four NVIDIA RTX 4090D GPUs (Ada Lovelace, 24GB). Specifically, the AutoProfiler rewriter targets Triton kernel generation. To isolate performance gains driven by profiling policy from baseline framework translation artifacts, each original PyTorch operator is first translated into a correctness-verified Triton seed prioritizing numerical fidelity over performance.

\subsection{Main Results}
Table~\ref{tab:main_results} summarizes the performance across different methods on KernelBench. 
AutoProfiler achieves the highest average speedup of {1.704$\times$}, outperforming the strongest baseline, CudaForge (1.570$\times$) by 8.5\%. 
Notably, the performance advantage of AutoProfiler is most pronounced on the challenging Level 3 task, where it elevates the speedup from 1.099$\times$ of CudaForge to {1.352$\times$}. 
This substantial gain demonstrates that kernel-conditioned profile selection scales effectively to complex optimization spaces, going beyond simple operator-level enhancements.

In terms of correctness, AutoProfiler attains the highest execution accuracy at {87.2\%}, outperforming other baselines by 2.4--8.8 percentage points. 
Crucially, these efficiency and accuracy gains are achieved with minimal compute overhead: AutoProfiler requires only one single LLM call per task, whereas CudaForge and KernelMem consume 2 calls, and KernelAgent requires 4 calls. 
These results substantiate that profiling-guided hardware feedback serves as an efficient, lightweight prior for LLM-driven kernel optimization, delivering superior optimization quality while markedly reducing generation costs.

\begin{table}[t]
\centering
\small
\caption{Main performance comparison on KernelBench. ``Speedup'' denotes the geometric mean speedup over correctly solved problems, while ``Accuracy'' denotes the proportion of problems solved correctly. ``LLM Calls/Task'' indicates the average number of LLM inferences required per optimization task.}
\label{tab:main_results}
\begin{tabular}{lcccccc}
\toprule
\multirow{2}{*}{Method} &
\multicolumn{4}{c}{Speedup~($\uparrow$)} &
\multirow{2}{*}{Avg. Accuracy (\%)} &
\multirow{2}{*}{LLM Calls/Task} \\
& {Level 1} & {Level 2} & {Level 3} & {Avg.}& & \\
\midrule
CudaForge    & 1.567$\times$ & \textbf{1.808}$\times$ & 1.099$\times$ & 1.570$\times$ & 78.4 & 2 \\
KernelMem    & 1.233$\times$ & 1.351$\times$ & 1.091$\times$ & 1.252$\times$ & 82.0 & 2 \\
KernelAgent  & 1.361$\times$ & 1.306$\times$ & 1.044$\times$ & 1.276$\times$ & 84.8 & 4 \\
AutoKernel   & 1.058$\times$ & 1.182$\times$ & 0.997$\times$ & 1.095$\times$ & 84.4 & 1 \\ \midrule
AutoProfiler & \textbf{1.810$\times$} & 1.775$\times$ & \textbf{1.352$\times$} &
\textbf{1.704$\times$} & \textbf{87.2} & \textbf{1} \\
\bottomrule
\end{tabular}
\end{table}

\subsection{Further Analysis}
\subsubsection{Comparison of Profiling Feedback Strategies}
We evaluate five distinct control conditions over a representative subset of 48 diverse kernels spanning across all difficulty levels: 
(1) \textit{Blind}: provides no profiling metadata; 
(2) \textit{Random}: attaches a randomly sampled profiling configuration; 
(3) \textit{Static}: uses a fixed, globally optimal profiling setup; 
(4) \textit{{AutoProfiler}}: dynamically assigns kernel-adaptive profiling trees; and 
(5) \textit{Oracle}: assigns the empirical upper-bound configuration for each individual kernel, which is unavailable during inference. 
All other components, including the rewriter, optimization pipeline, hardware, and sampling seeds, are kept identical.

As summarized in Table~\ref{tab:profiling}, profiling feedback consistently improves kernel optimization. 
Our \textit{{AutoProfiler}} achieves a 1.379$\times$ speedup, compared with 1.197$\times$ for \textit{Blind}, and also outperforms both \textit{Random} and \textit{Static}.
These results show that the benefit comes not only from providing profiling information, but from selecting feedback adaptively for each kernel. The remaining gap to \textit{Oracle} suggests further room for improving profiling selection.


\subsubsection{More Profiling Is Not Always Better}

We further investigate how the volume of profiling feedback influences downstream optimization efficacy. Starting from the learned 4-metric setup as our reference, we construct reduced variants by pruning metrics and expanded ones by sampling additional metrics from the NCU pool. As presented in Table~\ref{tab:quantity}, downstream performance exhibits a clear non-monotonic trend with respect to profiling metric quantity. Reducing feedback from four metrics to two drops the geomean speedup from 1.370$\times$ to 1.299$\times$. Conversely, over-saturating the context with six or eight metrics degrades performance further to 1.261$\times$ and 1.231$\times$, respectively. These results demonstrate that effective profiling requires judicious metric selection to maximize information density, avoiding both metric scarcity and redundancy.

\begin{table}[t]
    \centering
    \small
    \begin{minipage}[t]{0.4 \linewidth}
        \centering
        \caption{Effect of profiling information on downstream kernel optimization. }
        \label{tab:profiling}
        \vspace{2pt}
        \begin{tabular}{lc}
        \toprule
        {Condition} & { Speedup~($\uparrow$)} \\
        \midrule
        Blind  & 1.197 \\
        Random & 1.215 \\
        Static & 1.314 \\
        {AutoProfiler}   & \textbf{1.379} \\
        Oracle & 1.573 \\
        \bottomrule
        \end{tabular}
    \end{minipage}%
    \hfill 
    \begin{minipage}[t]{0.55 \linewidth}
        \centering
        \caption{Impact of profiling information quantity on downstream kernel optimization.}
        \label{tab:quantity}
        \vspace{2pt}
        \begin{tabular}{lcc}
        \toprule
        {\# Metrics} & {Regime} & {Speedup~($\uparrow$)} \\
        \midrule
        2 & Insufficiency & 1.299$\times$ \\
        \textbf{4} & \textbf{Reference Baseline} & \textbf{1.370$\times$} \\
        6 & Redundancy & 1.261$\times$ \\
        8 & Redundancy & 1.231$\times$ \\
        \midrule
        Static & -- & 1.219$\times$ \\
        \bottomrule
        \end{tabular}
    \end{minipage}
\end{table}

\subsubsection{Generalization Across LLM Backbones}
\label{sec:cross-rewriter}

To test whether the learned profiling policy captures intrinsic hardware requirements rather than overfitting to a specific LLM, we evaluate its transferability across model families. Specifically, we freeze the profiling tree trained on GPT-5.6-Sol and replace only the leaf kernel-rewriting LLM backbone with five distinct models, evaluating each on held-out benchmarks under the same pipeline.

As summarized in Table~\ref{tab:cross-rewriter}, the frozen profiling policy generalizes strongly across all backbones. When paired with Claude-Sonnet-5, the system achieves a peak geomean speedup of 1.47$\times$ and an execution accuracy of 91.7\%, outperforming the training time baseline. Furthermore, across backbones from diverse model families (\emph{e.g.,} Kimi-K3, DeepSeek-V4-Pro, and Qwen3-Coder-Plus), the profiling policy consistently guides effective kernel optimization. These results confirm that the learned profiling tree encodes hardware optimization priors that are independent of the underlying LLM backbone rather than model specific biases.

\begin{table}[h]
\centering
\small
\caption{Generalization performance across different LLM backbones. The profiling tree is constructed using GPT-5.6-Sol feedback and kept strictly frozen; only the leaf LLM is swapped at test time. Speedup denotes the geometric mean over PyTorch baselines on held-out kernels.}
\label{tab:cross-rewriter}
\begin{tabular}{llcccc}
\toprule
Rewriter & Model Family & Speedup~($\uparrow$) & Accuracy~($\uparrow$) & Time (s, $\downarrow$) & Tokens~($\downarrow$) \\
\midrule
Claude-Sonnet-5 & Anthropic       & 1.47$\times$ & 91.7 & 8.2  & 1,840 \\
Kimi-K3         & Moonshot        & 1.38$\times$ & 89.6 & 9.1  & 2,010 \\
GPT-5.6-Sol     & OpenAI    & 1.34$\times$ & 87.5 & 10.4 & 2,260 \\
DeepSeek-V4-Pro & DeepSeek        & 1.21$\times$ & 83.3 & 12.7 & 2,580 \\
Qwen3-Coder-Plus & Qwen            & 1.12$\times$ & 79.2 & 14.3 & 3,120 \\
\bottomrule
\end{tabular}
\end{table}

\subsubsection{Iterative Kernel Refinement}
We evaluate multi-turn optimization by recursively applying the frozen profiling tree over ten consecutive rounds, using the kernel produced in each step as the input for the next. 
Figure~\ref{fig:pareto} reports the optimization trajectory as a function of cumulative wall-clock time per sample, providing a direct comparison with CudaForge under actual optimization cost. 
AutoProfiler improves rapidly from 1.629$\times$ in the first round to 2.112$\times$ after seven rounds, while requiring substantially less time per iteration. 
In comparison, CudaForge reaches 1.846$\times$ after four rounds under a similar cumulative time budget. 
These results show that the learned profiling policy not only supports stable iterative refinement, 
but also enables a more favorable optimization-cost trade-off by avoiding repeated expensive profiling and LLM-based profiling decisions.



\subsection{Ablation Study}

\begin{figure}[t]
    \centering
    \small
    \begin{minipage}[c]{0.4\linewidth}
        \centering
        \includegraphics[width=\linewidth, keepaspectratio]{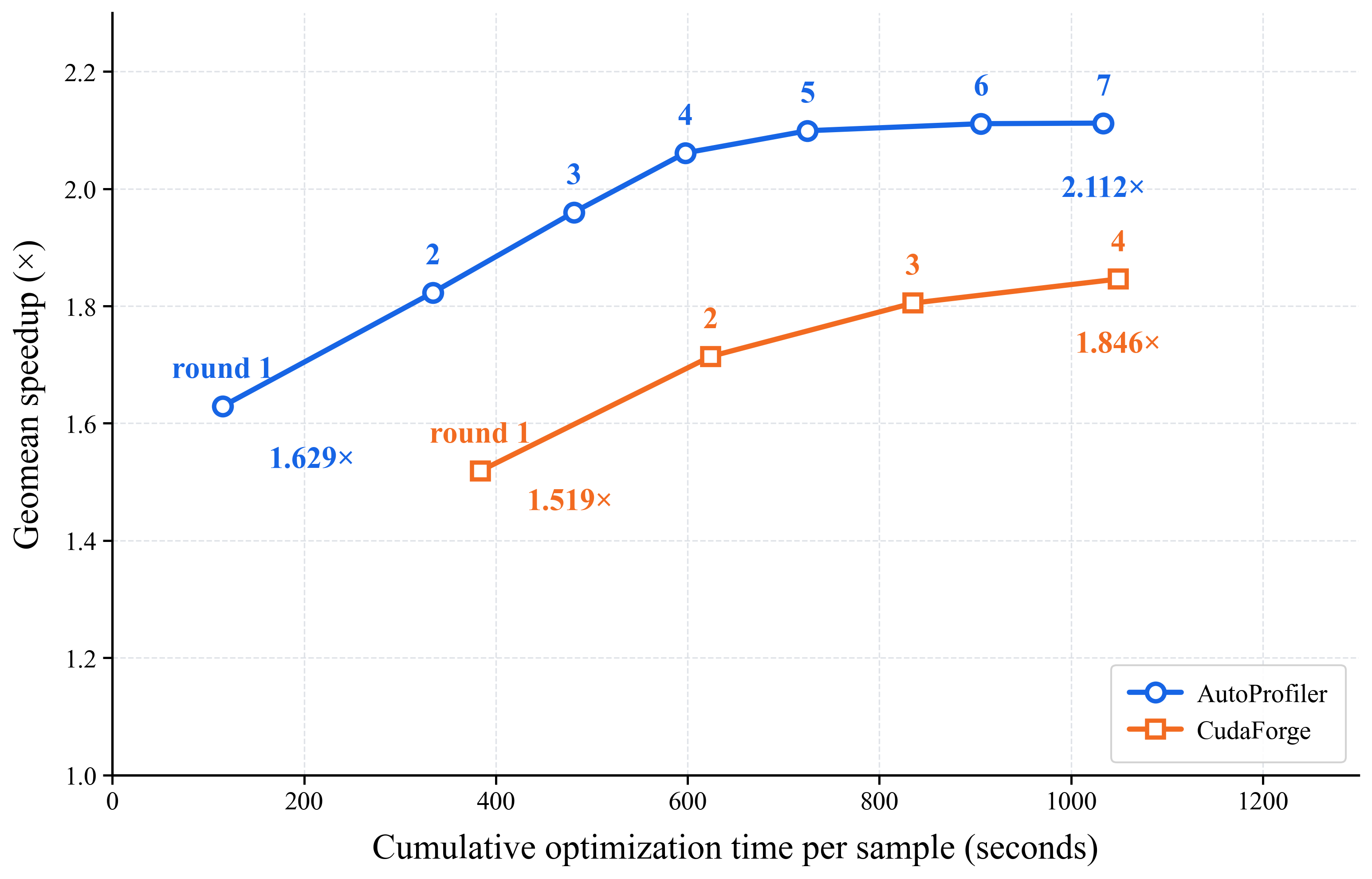}
        \caption{Comparison of optimization trajectories. Each circled marker represents an optimization round.}
        \label{fig:pareto}
    \end{minipage}%
    \hfill 
    \begin{minipage}[c]{0.55\linewidth}
        \centering
        \small
        \begin{tabular}{lccc}
        \toprule
        {Configuration} & {Depth} & {\# Leaves} & {Speedup~($\uparrow$)} \\
        \midrule
        {AutoProfiler}       & 11 & 58 & \textbf{1.61$\times$} \\
        w/o REWRITE                 &  2 &  4 & 1.47$\times$ \\
        w/o SPLIT                    &  7 & 37 & 1.53$\times$ \\
        \bottomrule
        \end{tabular}
        \captionof{table}{Ablation study on tree-construction operators.}
        \label{tab:ablation}
    \end{minipage}
\end{figure}

As summarized in Table~\ref{tab:ablation}, removing either operator causes a clear performance drop, demonstrating that both components are essential for optimal tree construction. Removing \textsc{Rewrite} drops the geomean speedup from {1.61$\times$} to {1.47$\times$}. 
In this configuration, while \textsc{Split} still divides kernels into separate branches, each leaf provides only basic metric feedback. Without the enriched profiling metrics, the downstream LLM lacks sufficient hardware context to guide fine-grained optimizations. Conversely, removing \textsc{Split} lowers performance to {1.53$\times$}. Without adaptive node splitting, the tree cannot separate kernels with distinct bottlenecks, forcing very different workloads to share the same profiling setup. The full policy achieves the peak speedup of {1.61$\times$} by combining both operators. These results confirm that \textsc{Split} groups kernels into specialized hardware regimes, while \textsc{Rewrite} ensures each regime receives the precise profiling details needed for LLM-driven code optimization.


\section{Conclusion}

In this paper, we present AutoProfiler, a novel feedback-driven framework that addresses the fundamental limitations of fixed hardware profiling in LLM-based GPU kernel optimization. 
Motivated by the observation that less profiling can yield better optimization, we shift from static metric collection to a profiling reasoning perspective, AutoProfiler learns a deterministic, kernel-conditioned profiling policy via iterative tree construction using REWRITE and SPLIT operations. Empirical results on KernelBench show that AutoProfiler outperforms traditional static profiling configurations and strong agentic baselines under constrained profiling budgets while incurring minimal LLM invocation overhead. Our findings suggest that hardware profiling should be treated as a learnable decision process, opening a promising direction for future research in AI-driven systems optimization.

\bibliography{iclr2027_conference}
\bibliographystyle{iclr2027_conference}

\appendix
\section{Appendix}

\subsection{Related Work}

\paratitle{Traditional GPU Kernel Optimization.}
Automated GPU kernel optimization has long relied on search-based compilation frameworks and hardware-aware cost models. Early systems like TVM \citep{chen2018tvm} and AutoTVM \citep{chen2018learning} introduced learned cost models to guide program space exploration. Ansor \citep{zheng2020ansor} expanded this search space via hierarchical program sampling, while MetaSchedule \citep{shao2022tensor} unified rule-based and search-based program transformations. Further extensions such as FlexTensor \citep{zheng2020flextensor} and TensorIR \citep{feng2023tensorir} specialized in tensor program abstractions. While effective, these compiler-based approaches operate on rigid, template-driven program spaces and treat hardware profiling primarily as passive static evaluation or fixed performance metrics, lacking the adaptive code reasoning capabilities offered by modern large language models.

\paragraph{LLM-driven Kernel Optimization and Agentic Frameworks.}
With the rise of LLMs, recent efforts leverage generative models for GPU kernel synthesis and iterative optimization. Benchmarks like KernelBench \citep{ouyang2025kernelbench} have formalized the evaluation of LLM-generated CUDA and Triton kernels. Building on this, agentic systems integrate execution and runtime feedback into iterative search loops. For instance, CudaForge \citep{zhang2025cudaforge} and AutoKernel \citep{jaber2026autokernel} apply iterative compilation and error-repair heuristics, whereas KernelAgent \citep{cheng2026kernelagent} and Astra \citep{wei2025astra} employ multi-agent orchestration for kernel refactoring. Additionally, Kevin \citep{baronio2026kevin} and CUDA Agent \citep{dai2026cudaagent} explore reinforcement learning and dynamic tool invocation to guide kernel synthesis. Despite their success, these frameworks treat hardware profiling as an invariant preprocessing step or a static feedback prompt. They overlook the fact that hardware profiling is inherently costly and that metric utility varies across heterogeneous GPU kernels.

\paragraph{Hardware Profiling: From Static Metric Collection to Profiling Reasoning.}
Hardware profiling tools such as NVIDIA Nsight Compute \citep{nvidia2023nsight} provide hundreds of microarchitectural performance counters, but full metric acquisition incurs non-trivial latency and context redundancy. Early performance diagnosis tools like PerfExpert \citep{burtscher2010perfexpert} used static heuristic trees to isolate performance bottlenecks. Recent LLM-based micro-profiling assistants, such as KernelPro \citep{gai2026optimizing}, selectively trigger profiling tools based on predefined rule sets. However, existing methods either collect a universal metric subset indiscriminately or rely on hand-crafted profiling rules. In contrast, our work introduces the paradigm of \emph{profiling reasoning}: we formulate hardware metric acquisition as a kernel-conditioned sequential decision process, explicitly learning a deterministic profiling policy driven directly by downstream code optimization rewards.

\subsection{Empirical Verification of Profiling-Utility Heterogeneity}\label{app:empirical}

To empirically validate the heterogeneity assumption underlying Proposition 1, we quantify the performance discrepancy between static metric selection and dynamic, per-kernel metric selection on a representative set of 128 kernels sampled from \textsc{KernelBench}.

\paragraph{Candidate Configurations and Profiling Setup.}
We construct four static candidate profiling configurations, denoted as $\mathcal{C} = \{ m_{\text{algo}}, m_{\text{mem}}, m_{\text{lat}}, m_{\text{comp}} \}$, designed to capture distinct, roofline-relevant execution characteristics:
\begin{itemize}[leftmargin=*]
    \item \textbf{Algorithm} ($m_{\text{algo}}$): Captures high-level algorithmic behavior and control-flow patterns.
    \item \textbf{Memory} ($m_{\text{mem}}$): Focuses on DRAM/SRAM memory bandwidth utilization and transaction efficiency.
    \item \textbf{Latency} ($m_{\text{lat}}$): Measures instruction pipeline stalls, execution latency, and warp scheduling overheads.
    \item \textbf{Compute} ($m_{\text{comp}}$): Tracks compute unit saturation, floating-point throughput, and ALU utilization.
\end{itemize}
All metric sets are fixed prior to evaluation. For each kernel $k \in \mathcal{K}$ and candidate configuration $m_i \in \mathcal{C}$, we extract the profiling information $I_i(k) = P(k, m_i)$ and invoke a frozen kernel rewriter $g(\cdot)$ under identical optimization, hardware, and random seed settings. We record the resulting speedup relative to the PyTorch reference implementation as $s_i(k) = s(k, m_i)$.

\paragraph{Static Benchmark vs. Per-Kernel Oracle.}
We compare the performance of the single best static configuration against an empirical per-kernel oracle. The optimal fixed configuration $m^*$ is chosen to maximize the geometric mean speedup across the benchmark:
\begin{equation}
m^* = \operatorname*{argmax}_{m_i \in \mathcal{C}} \exp \left( \frac{1}{|\mathcal{K}|} \sum_{k \in \mathcal{K}} \ln s_i(k) \right).
\end{equation}
Conversely, the per-kernel oracle dynamically selects the optimal configuration independently for each individual kernel:
\begin{equation}
s_{\text{oracle}}(k) = \max_{m_i \in \mathcal{C}} s_i(k).
\end{equation}
Kernels for which no metric configuration yields a speedup over the unoptimized reference implementation ($s_i(k) \le 1.0$) are tracked separately under non-improving cases.

\paragraph{Empirical Findings and Headroom Analysis.}
As detailed in Table~\ref{tab:empirical}, no static configuration consistently dominates across heterogeneous workloads. While the strongest fixed baseline Algorithm achieves a geometric mean speedup of $1.332\times$, the per-kernel oracle reaches $1.429\times$—delivering a 7.3\% relative performance improvement.

To assess statistical confidence, we conduct $1,000$ bootstrap resamples over the 128 evaluation kernels, establishing a 95\% confidence interval of $[1.045\times, 1.107\times]$ for the relative gain of the oracle over the static baseline. Furthermore, the optimal profiling configuration shifts substantially across workloads: Memory provides the highest speedup for 31 kernels, Latency for 27, Algorithm for 26, and Compute for 23. These empirical observations highlight two key insights:
\begin{enumerate}[leftmargin=*]
    \item \textbf{High Utility Heterogeneity:} The value of profiling metrics is fundamentally kernel-dependent, rendering any fixed, one-size-fits-all profiling configuration inherently suboptimal.
    \item \textbf{Substantial Optimization Headroom:} Tailoring hardware profiling signals to specific kernel bottlenecks yields meaningful execution gains, underscoring the need for an adaptive profiling policy like \textsc{AutoProfiler}.
\end{enumerate}

\begin{table}[h]
\centering
\caption{Empirical heterogeneity of profiling utility on 128 KernelBench kernels.
Each candidate configuration is evaluated with the same frozen kernel optimizer.The ``Best on \# Kernels'' column counts kernels for which the configuration
achieves the highest downstream speedup among the four candidates.
The oracle selects the best candidate independently for each kernel.}
\label{tab:profiling_heterogeneity}
\begin{tabular}{lcc}
\toprule
Profiling Configuration & Geomean Speedup & Best on \# Kernels \\
\midrule
Algorithm & 1.332$\times$ & 26 \\
Memory    & 1.277$\times$ & 31 \\
Latency   & 1.242$\times$ & 27 \\
Compute   & 1.239$\times$ & 23 \\
\midrule
Per-kernel Oracle & \textbf{1.429$\times$} & 128 \\
\bottomrule
\end{tabular}
\label{tab:empirical}
\end{table}

\end{document}